\documentclass[lettersize,journal]{IEEEtran}
\usepackage{amsmath,amsfonts}
\usepackage{algorithmic}
\usepackage{algorithm}
\usepackage{array}
\usepackage[caption=false,font=normalsize,labelfont=sf,textfont=sf]{subfig}
\usepackage{textcomp}
\usepackage{stfloats}
\usepackage{url}
\usepackage{verbatim}
\usepackage{graphicx}
\usepackage{cite}
\usepackage{booktabs}
\usepackage{xcolor}

\begin{document}

\title{Detecting Defective Wafers Via Modular Networks}

\makeatletter
\newcommand{\linebreakand}{%
  \end{@IEEEauthorhalign}
  \hfill\mbox{}\par
  \mbox{}\hfill\begin{@IEEEauthorhalign}
}

\makeatother
\author{Yifeng Zhang, Bryan Baker, Shi Chen, Chao Zhang, Yu Huang, Qi Zhao$^*$\thanks{$^*$ denotes equal authorship.}, Sthitie Bom$^*$}

\markboth{IEEE TRANSACTIONS ON SEMICONDUCTOR MANUFACTURING}%
{Shell \MakeLowercase{\textit{et al.}}: A Sample Article Using IEEEtran.cls for IEEE Journals}


\maketitle


\begin{abstract}
The growing availability of sensors within semiconductor manufacturing processes makes it feasible to detect defective wafers with data-driven models. 
Without directly measuring the quality of semiconductor devices, they capture the modalities between diverse sensor readings and can be used to predict key quality indicators (KQI, \textit{e.g.}, roughness, resistance) to detect faulty products, significantly reducing the capital and human cost in maintaining physical metrology steps. 
Nevertheless, existing models pay little attention to the correlations among different processes for diverse wafer products and commonly struggle with generalizability issues.
To enable generic fault detection, in this work, we propose a modular network (MN) trained using time series stage-wise datasets that embodies the structure of the  manufacturing process. It decomposes KQI prediction as a combination of stage modules to simulate compositional semiconductor manufacturing, universally enhancing faulty wafer detection among different wafer types and manufacturing processes. 
Extensive experiments demonstrate the usefulness of our approach, and shed light on how the compositional design provides an interpretable interface for more practical applications.
\end{abstract}



\section{Introduction}
\IEEEPARstart{I}{n} the Industry 4.0~\cite{zhou2015industry} era, there is a remarkable increase in the number of smart sensors deployed in the semiconductor industry~\cite{botezatu2016predicting, petrov2021ieee}. 
They continuously produce valuable data that has the potential to be utilized for tracking and optimizing the manufacturing process~\cite{fan2020defective} as in reflecting the products' \textit{key quality indicators} (KQIs) for detecting faulty wafers. Substantial cost savings in terms of capital and human resources can be gained by modeling \textit{hard-to-measure} KQIs (\textit{e.g.}, resistance, flatness, \textit{etc.}) from \textit{easy-to-measure} sensor readings (\textit{e.g.}, temperature, pressure, \textit{etc.})~\cite{sun2021survey}.

\begin{figure}
    \centering
    \includegraphics[width=\linewidth]{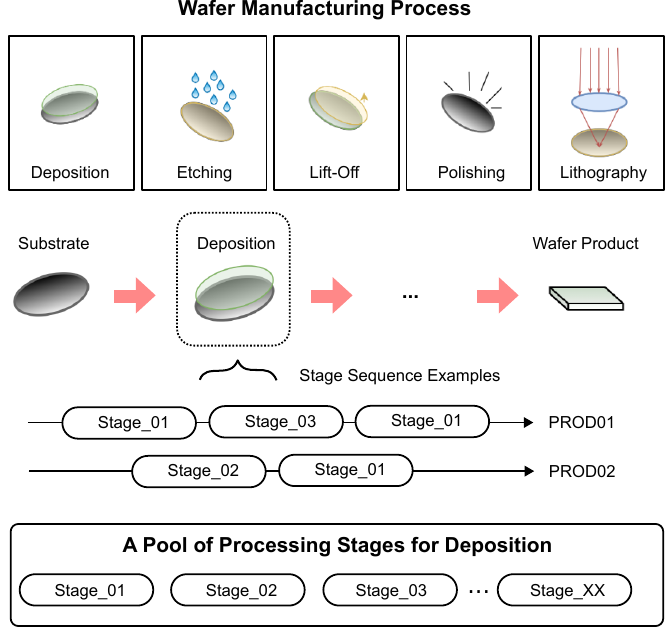}
    \caption{Illustration of the wafer manufacturing process.}
    \label{fig:teaser}
\end{figure}

{
Wafer products are processed using multiple manufacturing stages.  Each stage performs functions which can be reused depending on the product as well as the fabricated feature within a product.  For example, different wafer products (\textit{e.g.}, PROD01 and PROD02 in Figure~\ref{fig:teaser}) are produced as a combination of sharable manufacturing stages (\textit{e.g.}, Stage\_01).  To achieve high precision and recall for the KQI prediction, existing approaches focus on improving the structure of computational models and optimizing them with a variety of data splits. They prepare sensor data~\cite{8839832, 8634922, petrov2021ieee} in transaction-like format and train variants of state-of-the-art deep networks (\textit{e.g.}, autoencoders~\cite{kingma2013auto, sutskever2014sequence}, transformers~\cite{vaswani2017attention}, \textit{etc.}) to model the complexity in sensing modalities, \textit{e.g.}, multi-label stability~\cite{qian2021soft}, sensor imbalance~\cite{zhang2021soft}, and heterogeneous sensor dependency~\cite{huang2021grassnet}.
However, the black-boxed design of deep learning methods fails to reflect the complexity of semiconductor manufacturing resulting in inconsistent KQI prediction performances at multiple stages of diverse product lines.  Furthermore, the existing training data format using transaction-like data often omits crucial information about the manufacturing process (\textit{e.g.} the ordering of manufacturing stages).  Together, these limitations hinder the explicit modeling of intermediate procedures, creating substantial obstacles in identifying dependencies among different manufacturing processes and in cross-product generalization.
}

{
With an overarching goal to facilitate smart-sensing models with enhanced generalizability, we propose a modular network (MN) using state-wise sequence-like datasets to detect faulty semiconductor wafers.
The modular network centers around two major components, \textit{i.e.}, a list of functional prototypes which reflect base operations performed in the manufacturing pipeline, and a collection of stage modules that select relevant prototypes in the corresponding wafer manufacturing stage. This disentangles the diverse processing functions from different stages to naturally factor the entire manufacturing procedure for KQI predictions.
To further facilitate the training of our proposed network, we also generate corresponding datasets rich in quantity and annotations which explicitly illustrate the intermediate manufacturing procedures. In contrast to transaction-like soft sensing data~\cite{petrov2021ieee}, our datasets encapsulate the entire manufacturing stages of each product as well as the compositionality between stages. 
}

In summary, our major contributions are as follows:
\begin{enumerate}
    \item We are the first to propose a module-based method for key variable prediction tasks in the semiconductor wafer industry. It addresses the compositionality of manufacturing procedures and generalizes the model across wafer products and environment settings.
    {
    \item We propose a stage-wise sequence-like dataset that emphasizes compositonality of KQI predictions and is crucial for the learning of the modular network. 
    }
    \item We conduct extensive experiments over real-world manufacturing datasets with diverse settings to exhibit the usefulness of our proposed method.
\end{enumerate}

\section{Preliminaries}
The KQI prediction is core to the detection of faulty semiconductors, where a large body of studies focus on leveraging deep neural networks to capture the complex modalities between sensor data and KQIs~\cite{quirk2001semiconductor}. This section introduces the formulation of KQI prediction tasks and neural module networks use for this study.

\subsection{Key Quality Indicator Predictions}
{
To maintain product quality throughout the manufacturing process, smart sensors are installed in production tools to monitor wafers and to diagnos faulty stages.  The measurements at each processing stage, provide additional information for predicting/estimating hard-to-measure KQIs. Engineers also manually inspect sensor readings providing ground truth values (\textit{i.e.}, labels of KQIs) for training and evaluation. With high-dimensional sensor readings as the input and multiple KQI labels as the output, the key variable prediction can be formulated as a sequential classification task.}

{
Recent emphasis~\cite{kadlec2009data, petrov2021ieee} on detecting defective semiconductors has been placed on applying deep models over multiple tasks, such as with key variable indicator predictions~\cite{huang2021grassnet, zhang2021soft}, parameter control~\cite{birle2013fuzzy}, fault detection and diagnosis~\cite{jiang2021augmented}, along with other applications~\cite{abdin2021applying, qian2021soft}. They are commonly categorized into two main methods: transformer-based and graph-based. Transformer-based methods~\cite{qian2021soft, yella2021soft, zhang2021soft} leverage an encoder-decoder framework~\cite{sutskever2014sequence} with a multi-head attention mechanism~\cite{vaswani2017attention} to correlate sensor readings with wafer measurements, while graph-based methods~\cite{huang2021grassnet} learn to represent the latent correlation among sensors with both graph and graph attention propagation methods (\textit{e.g.} GAT~\cite{velivckovic2017graph}, GCN~\cite{defferrard2016convolutional},  GNN~\cite{scarselli2008graph}).
}

\begin{figure*}
    \centering
    \includegraphics[width=\linewidth]{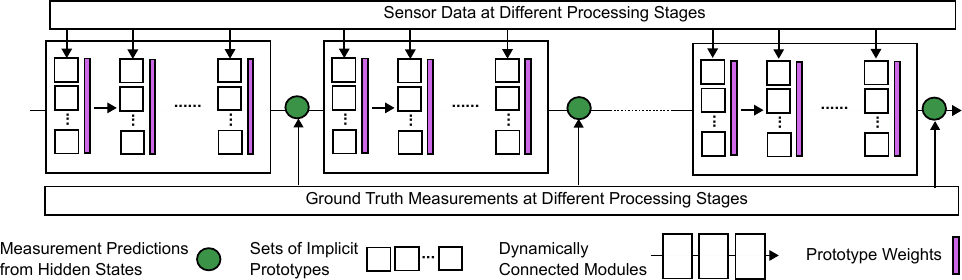}
    \caption{Overview of stage combinations in modeling wafer manufacturing. Our module-based network is composed of two major components, a collection of implicit prototypes which mimic base functions, and a set of stage modules which compose the manufacturing process. A weighted combination of implicit prototypes in each stage module is selected and recursively updated to mimic the physical functions performed within the manufacturing tools in the corresponding stage. }
    \label{fig:method_combine}
\end{figure*}

\subsection{Neural Module Networks}
In contrast to conventional black-box deep learning models~\cite{han2021transformer, lu2019vilbert}, Neural Module Networks~\cite{andreas2016neural, chen2021meta, hu2018explainable, hudson2019learning, shi2019explainable, zhang2021explicit, zhang2022query} were first introduced in the field of visual reasoning, which require answering questions based on visual input. They dynamically assemble modules into a network that is used to produce an answer in response to an input question. These modules play different roles within the network: querying the relevant knowledge by allocating or re-allocating attention to input features, recognizing the attended features, and performing numeric or logical operations. We adapt NMNs~\cite{shi2019explainable} traditionally used for vision tasks into our modular network to enable wafer manufacturing tasks with greater interpretability and generalizability. In doing so, designing modules that are closely linked to real-world stage procedures in wafer manufacturing tasks. These new modular networks differ from previous NMNs both conceptually and technically. 

Our method differs from existing deep models for faulty semiconductor detection in two major aspects. First, unlike approaches leveraging holistic black-box models directly correlating sensor readings to final measures, our modularized network provides a set of stage modules that align with the physical stage enabling the analysis of intermediate manufacturing procedures resulting in greater interpretability. Second, the stage modules are dynamically composed and better reflect the manufacturing processes of diverse products and environment settings resulting in significantly enhancing model generalizability.


\section{Methodology}
\label{sec:method}
Our proposed network addresses the complexity of processing stages by using a modular network (MN) outlined in Figure~\ref{fig:method_combine}. This network decomposes KQI predictions using three major thoughts: 
(A) a set of trained prototypes factorizing multiple base functions from their manufacturing procedures, 
(B) stage modules that predict measurements from sequential sensor readings within each stage,  
and (C) a joint objective function 
to supervise the learning of both the prototype and stage modules. Our proposed method is further refined in mimicking the manufacturing process by conducting a sequence of classification tasks. 

{
Let $\boldsymbol{X}=\{\boldsymbol{X}_1, \boldsymbol{X}_2, ..., \boldsymbol{X}_T\} \in R^{T\times D}$ represent $D$-dimensional sensor readings for $T$ consecutive stages and $\boldsymbol{L}=\{\boldsymbol{L}_1, \boldsymbol{L}_2, ..., \boldsymbol{L}_T\} \in R^{T\times K}$ represent the set of corresponding binary labels for $K$ KQIs (\textit{e.g.}, resistance, flatness, \textit{etc.}). 
The objective is to train a sequential model $\mathit{f}$ (\textit{i.e.}, LSTM~\cite{hochreiter1997long}) that is able to progressively predict KQIs, $\boldsymbol{L}_t$, at stage $t$ based on the set of corresponding sensor readings $\boldsymbol{X}_t$ and through hidden embeddings $\boldsymbol{h}_{t-1}$ obtained in previous stages. 
\begin{align}
    \boldsymbol{L}_t = f(\boldsymbol{X}_t, \boldsymbol{h}_{t-1})
\end{align}
}

In this section, we present details about the prototypes, stage modules, and the joint learning paradigm. 

\begin{figure*}
    \centering
    \includegraphics[width=\linewidth]{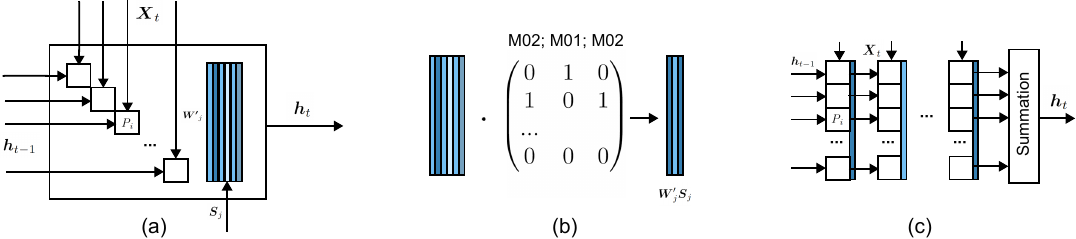}
    \caption{Illustration of Stage Module. (a) Stage Module Overview. (b) Select inter-prototype weights with mod selector $\boldsymbol{S}_j$. (c) Prototype Mappings at Different Mods.}
    \label{fig:method_stage}
\end{figure*}

\subsection{Implicit Prototypes}
\label{sec:prototype}
The key to compositionality, which enhances the generalizability and interpretability of KQI prediction/estimation, is the disentanglement of diverse base functions performed throughout the manufacturing process, \textit{e.g.}, heating a specific portion of airflow that can be shared among different processing stages. Due to the complexity of those fine-grained functions, it is impossible to model every physical function with a specific structure design. Hence, we propose to implicitly model them with a set of parameterized neural networks trained using multiple cross-measurement loss metrics (Discussed in Section~\ref{sec:method_learning}). 


To facilitate a tight connection between prototypes and physical base operations, we design the structure according to insights about how base functions participate in the manufacturing process. Conceptually, each base function impacts a portion of the wafer or environmental conditions. For example, changing the temperature or chamber pressure at a specific position within the manufacturing process is expected to influence a subset of KQIs due to the resulting change in sensor readings. A desirable prototype should be able to focus on relevant sensor inputs and conduct mappings based on the prototype index $i$, sensor inputs $\boldsymbol{X}_t$, and previous hidden states $\boldsymbol{h}_{t-1}$. Therefore, we simulate the set of prototypes using multiple-layer perceptrons and sensor masks. Specifically, each prototype $\textit{P}_i$ is parameterized by a trainable sensor mask $\boldsymbol{M}_i$ and intra-prototype weight $\boldsymbol{W}_i$. Given the sensor input $\boldsymbol{X}_t$ at stage $t$ and the previously hidden state embeddings $\boldsymbol{h}_{t-1}$, prototype $i$ mimics the operation by projecting the concatenation of masked sensor inputs with the hidden states into an output embedding $\boldsymbol{p}_{ti}$,
\begin{align}
    \mathit{P}_i (\boldsymbol{X}_t, \boldsymbol{h}_{t-1}) = \tanh(\boldsymbol{W}_i (\boldsymbol{X}_t \circ \boldsymbol{M}_i || \boldsymbol{h}_{t-1})),
\end{align}
where $||$ and $\circ$ are the concatenation operation and the hadamard multiplication, respectively. 

\subsection{Stage Module}
\label{sec:method_stage_module}
Aiming to enable KQI predictions with greater interpretability (\textit{e.g.}, fault diagnosis), we introduce a pool of stage modules that explicitly compose the manufacturing procedure. 
{
Ideally, a processing stage in a semiconductor wafer plant concurrently performs different combinations of functions~\cite{petrov2021ieee} repeatedly under the control of pre-programmed routines or recipes. As illustrated in Figure~\ref{fig:method_stage}(b), a stage goes through three repetitions of functions (\textit{i.e.}, named as mod with labels of MOD02, MOD04, MOD02) to produce the product. 
Assuming $j$ is the index of the module type, we construct the $j$-th module as a sequence of mappings to prototypes $\mathit{P}$, whose weights at different mods are controlled by inter-prototype weights $\boldsymbol{W'}_j$ and mod selector mask $\boldsymbol{S}_j$. Take Figure~\ref{fig:method_stage}(a) for example, the inter-prototype weights $\boldsymbol{W'}_j$ is a $I\times M$ matrix, where $I$ is the total number prototypes and $M$ is pre-set number of mod types based on the manufacturing process. Each column in the matrix refers to the inter-prototype weights for the corresponding mod. To obtain the inter-prototype weights, we project the concatenation of sensor readings $\boldsymbol{X}_t$ at $t$-th stage and the hidden states $\boldsymbol{h}_{t-1}$
\begin{align}
    \boldsymbol{W}'_j &= \mathit{f}_j(\boldsymbol{X}_t || \boldsymbol{h}_{t-1}),
\end{align}
where $\mathit{f}_j(\cdot)$ is a MLP with trainable parameters for the $j$-th type of stage.
}

{
After obtaining the inter-prototype weights for all possible mods, we generate the mod selector mask $\boldsymbol{S}_j$ by converting the information of mod sequences from the data using one-hot encoding. 
As shown in Figure~\ref{fig:method_stage}(c), with each column of mod selector mask $\boldsymbol{S}_j$ specifying the mod type, the output hidden stages $\boldsymbol{h}_t$ is computed by sequentially combining a weighted combination of prototypes for multiple mods,
\begin{align}
    \boldsymbol{h}_{t} &= \tanh(\boldsymbol{W}'_j\boldsymbol{S}_j\circ\boldsymbol{P}(\boldsymbol{X}_t, \boldsymbol{h}_{t-1}))
\end{align}
The hidden stage can further be fed into an MLP-based multi-label classifier $\mathit{f}_c$ to produce the binary KQI labels,
\begin{align}
    \boldsymbol{L}_t &= \mathit{f}_c (\boldsymbol{h}_t).
\end{align}
}



\subsection{Learning Prototypes and Stages}
\label{sec:method_learning}
To learn the sensor masks $\boldsymbol{M}_i$, intra-prototype weights $\boldsymbol{W}_i$, and the inter-prototype weights $\boldsymbol{W}'_j$ for stage modules, we dynamically compose the modules with the process stage combinations and train the models and prototypes in an end-to-end manner. The loss objectives consist of three major metrics, including (1) the measurement accuracy loss, (2) the prototype proximity loss, and (3) the prototype distinction regularization loss.

{
The \textbf{measurement accuracy loss}, $l_1$, estimates the cross-entropy between the sequence of ground-truth KQI predictions $\boldsymbol{L}_t$ and our predicted measurements $\boldsymbol{\hat{L}}_t$,
\begin{align}
    l_1 = \sum_{t=1}^{T}\sum_{i=1}^{K} L_{ti}\log L_{ti}+(1-L_{ti})\log(1-\hat{L}_{ti}),
\end{align}
where $\hat{L}_{ti}$ and $L_{ti}$ refer to the $i$-th predicted and ground truth KQI among a total of $K$ KQIs at the $t$-th stage, respectively.
}

{
The \textbf{prototype proximity loss}, $l_2$, takes into account the closeness of diverse prototypes over different KQI combinations. 
By re-encoding all the possible KQI combinations as a list of $K$ classes, our $I$ prototypes are distributed into $K$ classes equally. Prototypes within each class are expected to have close proximity in terms of margin-based or distance-based metrics. Therefore, we adopt from CPT~\cite{yang2018robust} the margin-based classification loss $l_\mathit{MCL}$ and distance-based cross entropy loss $l_\mathit{DCE}$ as prototype proximity loss
\begin{align}
    l_2 = l_\mathit{DCE} + l_\mathit{MCL}.
\end{align}
}

The \textbf{prototype distinction regularization loss}, $l_3$ limits the similarity among different prototypes and hence encourages prototypes to function on different portions of sensor readings. Given the intra-prototype weights $\boldsymbol{W}_i$ and the sensor masks $\boldsymbol{M}_i$, the prototype distinction regularization is found by computing the average cosine similarity between any two prototypes,
\begin{align}
    l_3 = -\sum_i \sum_j [\mathit{cos}(\boldsymbol{W}_i, \boldsymbol{W}_j) + \mathit{cos}(\boldsymbol{M}_i, \boldsymbol{M}_j)].
\end{align} 


%

\section{Dataset Formatting}
\label{sec:data}
{
For a comprehensive evaluation of the proposed method, we generate multiple data splits that emphasize different product or environment settings. In this section we detail the data schema, its generation paradigm, and the handling of missing entries. The data are created using the Seagate IEEE BigData Cup 2021~\cite{yella2021soft} public data source.
}

\begin{table}[]
    \centering
    \caption{The schema of raw data transactions.}
    \begin{tabular}{c|c }
    \toprule
        Category &  Sample Variables \\
    \midrule
        Meta Information & Process, Step, Stage, \textit{etc.} \\
        Manufacturing Program & Recipe, Tool, \textit{etc.}       \\
        Sensor Readings &  Temperature, Duration, Angle, \textit{etc.}  \\
        Measurements &  Thickness, Resistance, Flatness, \textit{etc.}     \\
    \bottomrule
    \end{tabular}
    \label{tab:data_schema}
\end{table}

{
\subsection{Schema Specifications.} Each sensor data transaction consists of four major components: the meta information, the manufacturing program, the sensor readings, and the measurements. As can be seen from the example in Table~\ref{tab:data_schema}, the meta information records the specific procedure (\textit{i.e.},  the name of the process, step, stage, and mod) in which the data is collected. The manufacturing program records the controlled parameters, such as the product type, the corresponding recipes, and tools.  The transactions also cover sensor readings and measurements, which document easy-to-measure environmental conditions, as well as the pass-fail status of certain semiconductor KQIs. 
}

{
\subsection{Data Generation Paradigm.} The transaction-like format of the sensor data enables vital information for the manufacturing procedures to be recorded. To model the temporal correlation among those transactions, raw data are grouped according to the identifier of the product and then ordered sequentially by transaction time. Specifically, for all the transactions belonging to a specific wafer, we first merge the measurements at the same procedure (\textit{i.e.}, with the same process/step/stage name), and then append them into a sequence ordered by ascending timestamp. For better training stability, we generate these sequences by moving a fixed-length window (\textit{i.e.}, window size $w$) over the initial sequence, alleviating the variability of raw data.
}

\subsection{Handling Missing Entries.}
In analyzing the data a number of missing entries are found in the sensor data, this is the result of: (1) environmental issues during manufacturing that randomly produce missing data, and (2) absence of sensors in different manufacturing tools that systematically provide missing data over a specific set of columns. We propose to handle these two types of missing data in different ways. For missing data due to environment issues, we adopt KNN methods~\cite{peterson2009k} to find a set of nearest neighbors describing similar manufacturing conditions and then leverage these to impute the missing entries. 
For the second type of missing data, we choose to handle them using a masking mechanism to lower the significance of these missing readings since absent sensors cannot provide information about the missing attributes.

\section{Experimental Results}
To fully evaluate the usefulness of our proposed approach, we conduct multiple experiments using variable settings for key prediction tasks.
In the following sections we present 
the implementation details (Section~\ref{sec:impl}), quantitative results (Section~\ref{sec:cs}) and ablation analysis (Section~\ref{sec:ab}) to demonstrate the capability of our method. 
 Discussions about interpretability are found in Section~\ref{sec:is}.


\subsection{Implementation Details}
\label{sec:impl}
{
\subsubsection{Experimental paradigm}
We evaluate our proposed method on both standard and generalized settings aimed at comprehensively evaluating these models’ performance in diverse scenarios. For the standard setting, models are evaluated on 11 sub-tasks (\textit{i.e.}, S1-S11) of the P1 toolset using the transaction-format dataset (\textit{i.e.}, Seagate IEEE BigData Cup 2021~\cite{yella2021soft}) that tests how the model responds to KQI predictions of a single manufacturing procedure. 
In contrast, for the generalized setting, we train and evaluate models using two data splits extracted from our proposed sequences. These training and evaluation splits contain sequences with nonidentical pools of a specific manufacturing program, \textit{e.g.}, product type, and product groups.
We adopt industry practice in reporting the AUC~\cite{bradley1997use} for both settings and in doing so track the overall performance of our approach.
For more generalizability tests involving other variables from the manufacturing program, please refer to the supplementary materials.
}

\subsubsection{Compared Models.}
To enable a comprehensive evaluation of our method, we compare our results with three state-of-the-art deep models applied to wafer fault detection.  These include the conventional LSTM~\cite{hochreiter1997long} as well as state-of-the-art transformer-based approaches (\textit{i.e.}, Transformer and Conformer models). All models are tested using both standard and generalized settings by adjusting the sequence length (\textit{i.e}, $1$ for standard settings and $w$ for generalized settings). 
For detailed hyperparameters used in the model, please refer to the supplementary materials.

\subsection{Quantitative Results}
\label{sec:cs}
To comprehensively evaluate the effectiveness of our proposed method, we compare our proposed network with the state-of-the-art methods in both standard and generalizing settings.

\begin{table}[!h]
    \centering
    \caption{Comparison with state-of-the-art using standard settings.}
    \begin{tabular}{c|c c c c}
        \toprule
        Tasks &  LSTM & Transformer & Conformer & Ours\\
        \midrule
         S1 & 0.61$\pm$0.01 & 0.70$\pm$0.12             & 0.72$\pm$0.08             & \textbf{0.75}$\pm$0.07 \\
         S2 & 0.43$\pm$0.05 & 0.60$\pm$0.18             & \textbf{0.72}$\pm$0.08    & 0.63$\pm$0.12 \\
         S3 & 0.48$\pm$0.04 & \textbf{0.86}$\pm$0.01    & 0.81$\pm$0.00             & 0.83$\pm$0.02 \\
         S4 & 0.49$\pm$0.01 & \textbf{0.91}$\pm$0.01    & 0.88$\pm$0.02             & \textbf{0.91}$\pm$0.04 \\
         S5 & 0.44$\pm$0.02 & 0.55$\pm$0.03             & \textbf{0.67}$\pm$0.03    & \textbf{0.67}$\pm$0.01 \\
         S6 & 0.53$\pm$0.02 & 0.53$\pm$0.05             & \textbf{0.64}$\pm$0.04    & \textbf{0.64}$\pm$0.04 \\
         S7 & 0.51$\pm$0.03 & 0.64$\pm$0.02             & \textbf{0.65}$\pm$0.01    & \textbf{0.65}$\pm$0.03 \\
         S8 & 0.38$\pm$0.02 & \textbf{0.82}$\pm$0.03    & 0.58$\pm$0.00             & 0.80$\pm$0.04 \\
         S9 & 0.63$\pm$0.01 &0.71$\pm$0.09              & 0.75$\pm$0.06             & \textbf{0.76}$\pm$0.04 \\
         S10 & 0.46$\pm$0.01&0.92$\pm$0.03              & \textbf{0.94}$\pm$0.00    & 0.87$\pm$0.09 \\
         S11 & 0.62$\pm$0.05&\textbf{0.89}$\pm$0.01     & 0.88$\pm$0.02             & 0.87$\pm$0.04 \\
        \bottomrule
    \end{tabular}
    \label{tab:cs}
\end{table}

{
\subsubsection{Standard Settings}
Table~\ref{tab:cs} lists the results of KQI predictions for 11 tasks. Overall, our method achieves the highest AUC on 6 out of 11 tasks, demonstrating better stability against different scenarios. Compared with other state-of-the-art methods having low AUC on specific tasks, (\textit{e.g.}, LSTM, Conformer on S8, and Transformer on S5), our method demonstrates an improvement in performance stability across diverse tasks. These results suggest that our method is capable of generalizing complex scenarios by addressing compositionality using latent prototypes.
It indicates that our modular network 
is capable of helping engineers locate product quality issues for solutions promptly.
}

{
\subsubsection{Generalized Settings}
\label{sec:gs}
To further evaluate the model's generalizability to different manufacturing procedures, we create multiple data splits with non-identical pools of product types and product groups. Each product type denotes wafers that share the same or similar semiconductor structure and design. The product group denotes a collection of product types with similar physical semiconductor functionalities. 
}

\begin{table}[!t]
    \centering
    \caption{Comparison with state-of-the-art using generalzed settings. (product type).}
    \begin{tabular}{c|c c c c}
       \toprule
       KQIs & LSTM & Transformer & Conformer & Ours \\
       \midrule
       M1 & 0.64$\pm$0.10  &0.71$\pm$0.01	           &0.77$\pm$0.08	&\textbf{0.82}$\pm$0.03     \\
       M2 & 0.55$\pm$0.03	&0.77$\pm$0.02	           &0.74$\pm$0.04	&\textbf{0.83}$\pm$0.03     \\
       M3 & 0.51$\pm$0.03	&0.63$\pm$0.01	           &\textbf{0.78}$\pm$0.02	&0.72$\pm$0.05      \\
       M4 & 0.82$\pm$0.08	&0.87$\pm$0.05	           &0.90$\pm$0.01	   &\textbf{0.92}$\pm$0.01     \\
       M5 & 0.76$\pm$0.03	&\textbf{0.84}$\pm$0.02	   &0.81$\pm$0.01	&0.76$\pm$0.08      \\
       M6 & 0.64$\pm$0.05	&0.90$\pm$0.01	           &0.82$\pm$0.02	   &\textbf{0.96}$\pm$0.02     \\
       M7 & 0.52$\pm$0.03	&\textbf{0.93}$\pm$0.03	   &0.86$\pm$0.00	&0.86$\pm$0.00      \\
       M8 & 0.56$\pm$0.02	&0.77$\pm$0.07	           &\textbf{0.87}$\pm$0.04	&0.85$\pm$0.02      \\
       M9 & 0.63$\pm$0.04	&0.62$\pm$0.08	           &0.73$\pm$0.04	&\textbf{0.76}$\pm$0.01     \\
       M10 & 0.69$\pm$0.11	&0.73$\pm$0.02  	       &0.84$\pm$0.02	&\textbf{0.87}$\pm$0.02     \\
       \bottomrule
    \end{tabular}
    \label{tab:gsp}
\end{table}

{
In Table~\ref{tab:gsp} we evaluate whether our method is capable of generalizing across diverse product types using the AUC metric.
Our approach outperforms the state-of-the-art methods on 6 out of 10 measurements, suggesting enhanced generalizability to the manufacturing process across product types.  It is also noteworthy that our methods achieve the lowest average variance. These results highlight the importance of taking a modular based model approach for KQI prediction tasks.
}

\begin{table}[!h]
    \centering
    \caption{Comparisons of the state-of-the-art and our method using generalzed settings (product group).}
    \begin{tabular}{c|c c c c}
       \toprule
       KQIs & LSTM & Transformer & Conformer & Ours \\
       \midrule
       M1 & 0.61$\pm$0.05  &0.68$\pm$0.04	           &0.71$\pm$0.03	&\textbf{0.79}$\pm$0.03  \\
       M2 & 0.52$\pm$0.06	&0.72$\pm$0.02	           &0.71$\pm$0.02	&\textbf{0.78}$\pm$0.01  \\
       M3 & 0.54$\pm$0.02	&0.66$\pm$0.03	           &0.72$\pm$0.10	&\textbf{0.74}$\pm$0.02   \\
       M4 & 0.66$\pm$0.04	&0.77$\pm$0.05	           &0.82$\pm$0.02	   &\textbf{0.86}$\pm$0.02     \\
       M5 & 0.70$\pm$0.07	&0.77$\pm$0.02	           &0.79$\pm$0.02	&\textbf{0.80}$\pm$0.03      \\
       M6 & 0.67$\pm$0.02	&0.73$\pm$0.01	           &0.77$\pm$0.02	   &\textbf{0.79}$\pm$0.02     \\
       M7 & 0.56$\pm$0.04	&0.80$\pm$0.02	           &\textbf{0.82}$\pm$0.01	&\textbf{0.82}$\pm$0.01      \\
       M8 & 0.54$\pm$0.03	&0.72$\pm$0.03	           &0.80$\pm$0.02	&\textbf{0.81}$\pm$0.01      \\
       M9 & 0.61$\pm$0.02	&0.64$\pm$0.05	           &\textbf{0.70}$\pm$0.04	&0.68$\pm$0.06     \\
       M10 & 0.62$\pm$0.06	&\textbf{0.71}$\pm$0.02  	       &0.68$\pm$0.03	&0.68$\pm$0.04     \\
       \bottomrule
    \end{tabular}
    \label{tab:ctf}
\end{table}

{
In manufacturing, different products (\textit{i,e.}, wafer) can have similar functions and therefore require similar procedures to produce. To extensively evaluate our model's generalizability across different procedures we utilize product groups to denote a list of similar product types combined or used interchangeably in hardware (\textit{e.g.}, a hard disk). Table~\ref{tab:ctf} demonstrates the AUC results of our network trained and evaluated across product groups. Compared to Table~\ref{tab:gsp}, all models demonstrate a drop in AUC performance for most of the KQIs. This indicates greater challenges for generalizing across product groups. Despite the elevated difficulty, our modular network still outperforms the state-of-the-art models (\textit{i.e.}, 8 out of 10), again stressing the essential role modular based modeling has in improving generalizability.
}


\begin{table}[!h]
    \centering
    \caption{Average AUC of different combinations of losses.}
    \begin{tabular}{c c c | c c}
    \toprule
        $l_1$ & $l_2$ & $l_3$ & Product & Product Group \\
    \midrule
        \checkmark & \checkmark & \checkmark & \textbf{0.84}$\pm$0.03 & \textbf{0.78}$\pm$0.03 \\
        - & \checkmark & \checkmark &  0.80$\pm$0.03  & 0.70$\pm$0.02\\
        \checkmark & - & \checkmark &  0.78$\pm$0.01  & 0.72$\pm$0.02\\
        \checkmark & \checkmark & - &  0.81$\pm$0.02  & 0.68$\pm$0.01\\
        \checkmark & - & -  & 0.76$\pm$0.03           & 0.60$\pm$0.08\\
        - & \checkmark & - &  0.72$\pm$0.02           & 0.60$\pm$0.06\\
        - & - & \checkmark &  0.73$\pm$0.06           & 0.64$\pm$0.01\\
    \bottomrule
    \end{tabular}
    \label{tab:losses}
\end{table}

\subsection{Ablation Analysis}
\label{sec:ab}
{
In this section, we ablate the crucial components to demonstrate how they address compositionality and contribute to KQI prediction. For simplicity, we experiment with generalized settings and report the average AUC across 10 selected KQIs.
}

{
\subsubsection{Impacts of Losses}
To evaluate how the three losses in our objective contribute to the model's performance, we conduct experiments using different combinations of losses and record the AUC results in Table~\ref{tab:losses}. This suggests that the combination of all three losses achieves the top performance, highlighting the necessity in taking into account the overall classification accuracy, the cross-measurement closeness prototypes, along with the distinction across prototypes in enhancing model performance. The absence of certain losses results in larger performance drops suggesting their crucial role in enhancing the generalizability across scenarios. 
}

\begin{table}[!h]
    \centering
    \caption{Average AUC of different model constructional components.}
    \begin{tabular}{c c | c c}
    \toprule
        Prototypes & Stage Modules & Product & Product Group \\
    \midrule
        \checkmark & \checkmark & \textbf{0.84}$\pm$0.03 & \textbf{0.78}$\pm$0.03 \\
        \checkmark & - &  0.68$\pm$0.03 & 0.66$\pm$0.04 \\
        - & \checkmark &  0.72$\pm$0.02 & 0.68$\pm$0.03 \\
        - & - & 0.66$\pm$0.06 & 0.62$\pm$0.04 \\
    \bottomrule
    \end{tabular}
    \label{tab:abla_com}
\end{table}

{
\subsubsection{Impacts of Constructional Components}
Our method takes advantage of two constructional components: 1) prototypes and 2) stage modules, to predict KQIs by using the compositionality of their manufacturing procedures. We ablate their functionalities and report the results in Table~\ref{tab:abla_com}. Overall, the introduction of each component achieves better performance compared to the baseline, suggesting their necessity in addressing model compositionality. 
They also demonstrate cooperative roles by proving the best performance when combined.
These results highlight the importance of using our method for these KQI predictions.}

\begin{table}[!h]
    \centering
    \caption{Inter Group Weight Correlations. Numbers in the upper triangle are ground truth inter-group product similarity based on group overlapping, while those in the lower triangle are cosine similarity of corresponding stages' inter-prototype weights.}
    \begin{tabular}{c|c c c c c}
         \toprule
         & PROD15 & PROD20 & PROD21 & PROD41 & PROD63  \\
         \midrule
         PROD15 &  1 & 0.51 & 0.37 & 0.54 & 0.39\\
         PROD20 &  0.47 & 1 & 0.71 & 0.58 & 0.42\\
         PROD21 &  0.32 & 0.63 & 1 & 0.77 & 0.47\\
         PROD41 &  0.46 & 0.54 & 0.67 & 1 & 0.62\\
         PROD63 &  0.38 & 0.65 & 0.43 & 0.51 & 1\\
         \midrule
         \multicolumn{6}{c}{Pearson (statistic=0.71, pvalue=0.022)} \\
         \bottomrule
    \end{tabular}
    \label{tab:ca_p1}
\end{table}

\begin{table}[!h]
    \centering
    \caption{Intra Group Weight Correlations. Not all groups are listed.}
    \begin{tabular}{c|c c c c c}
         \toprule
         & PROD20 & PROD21 & PROD41 & PROD63 \\
         \midrule
         Attention Similarity & 0.79 & 0.72 & 0.69 & 0.74 \\
         Product Similarity & 0.65 & 0.61 & 0.59 & 0.6\\
         \midrule
         \multicolumn{5}{c}{Pearson (statistic=0.75, pvalue=0.013)} \\
         \bottomrule
    \end{tabular}
    \label{tab:ca_p2}
\end{table}

\subsection{Interpretability Discussion}
\label{sec:is}
Compared with other black-box deep models, our modular network provides a transparent interface to reveal the role different stages have in the manufacturing process. Here, we provide quantitative analysis to validate the relationships between inter/intra-group products and prototype weights of stage modules. Specifically, products within the same group (\textit{i.e.}, intra-group) are expected to share similar functionalities compared to inter-group products. We further discuss the possible industry applications.

For each product group, we compute the average cosine similarities of prototype weight vectors for both inter-group and intra-group products. Similar products are found in similar groups. We also approximate the ground-truth similarity across product groups based on if their group members occur in other groups. Table~\ref{tab:ca_p1} and Table~\ref{tab:ca_p2} demonstrate inter/intra-group attention similarity and product similarity across product groups, respectively. It should be noted that each group is selected to represent a main product category. Results using Pearson correlations show that, in both cases, the cosine similarity of attention vectors aligns with product similarity with high confidence, suggesting the potential to leverage learned stage modules in analyzing or diagnosing possible faults. 

{
With correlations between our designed stage modules and the physical manufacturing stages, our method enables several potential applications, including faulty factor reasoning and fast prototyping. For instance, By outputting the KQI predictions for every intermediate stage, our method provides extra clues about the production, illustrating when the current product under certain recipes is likely to be defective. Further, our method has the potential to support prototyping of new product KQIs using a combination of pretrained previous manufacturing processes. Both applications play a crucial role in accelerating the diagnosis and development of semiconductors. }

\section{Conclusion}
This paper proposes a novel module-based network for soft sensing tasks. It addresses the compositionality of key variable prediction tasks using a pool of prototypes that disentangle base functions performed on wafers, as well as a set of stage modules that dynamically compose the manufacturing process. Results on multiple splits validate that our proposed method is capable of generalizing to both conventional settings and generalizability settings with a variety of changing factors (\textit{e.g.}, product type, product group). Additional interpretability analysis also demonstrates the potential of our method to be applied as a transparent platform for several applications, including possible fault diagnosis. We hope our work will be useful for the future development of general and interpretable soft sensing models.

\bibliographystyle{IEEEtran}
\bibliography{journalmain}

\newpage


\vfill

\end{document}


\title{Detecting Defective Wafers Via Neural Module Networks (Supplementary Materials)}


\markboth{IEEE TRANSACTIONS ON SEMICONDUCTOR MANUFACTURING}%
{Shell \MakeLowercase{\textit{et al.}}: A Sample Article Using IEEEtran.cls for IEEE Journals}


\maketitle

In the main paper, we have presented the data contribution and modular network that address the compositionality in KQI predictions. The supplementary materials provide additional experimental results and implementation details of our proposed work.
\begin{itemize}
    \item Section~\ref{sec:hyper} provides ablation analyses of the hyperparameters used in dataset generation and modular network.
    \item Section~\ref{sec:add_e} provides additional experimental results in generalizing settings, including the tests in terms of time and data scales.
\end{itemize}

\section{Ablation of Hyperparameters}
\label{sec:hyper}
\begin{table}[!h]
    \centering
    \caption{Precision of different approaches to handle measurement issues. [a, b] denotes the label assigned to the missing entry in the hardcoding approach.}
    \begin{tabular}{c|c c c c | c | c}
       \toprule
       KQIs & [1,1] & [0,0] & [0,1] & [1,0] & Hard Loss & Soft Loss \\
       \midrule
       M1	& 0.83 & 0.84 & 0.83 & 0.83 & \textbf{0.85} & \textbf{0.85} \\
       M2	& 0.85 & 0.85 & 0.81 & 0.85 & \textbf{0.86} & \textbf{0.86} \\
       M3 & 0.86 & 0.85 & 0.80 & 0.84 & \textbf{0.88} & \textbf{0.88}\\
       M4	&\textbf{0.85} & 0.84 & 0.82 & 0.83 & \textbf{0.85} & \textbf{0.85}\\
       M5 &0.84 & 0.83 & 0.81 & 0.83 & \textbf{0.85} & \textbf{0.85} \\
       M6	&0.88 & 0.85 & 0.82 & 0.84 & 0.90 & \textbf{0.91} \\
       M7	&0.85 & 0.85 & 0.81 & 0.81 & 0.86 & \textbf{0.87}  \\
       M8	&0.83 & 0.83 & 0.81 & 0.83 & \textbf{0.85} & 0.84\\
       M9 &0.71 & 0.66 & 0.64 & 0.68 & 0.72 & \textbf{0.75}\\
       M10	&0.90 & 0.87 & 0.83 & 0.86 & \textbf{0.91} & \textbf{0.91}\\
       M11	&0.81 & 0.81 & 0.78 & 0.81 & \textbf{0.83} & \textbf{0.83}\\
       M12	&0.82 & 0.81 & 0.78 & 0.81 & 0.82 & \textbf{0.84}\\
       M13	&\textbf{0.86} & 0.82 & 0.82 & 0.82 & 0.85 & 0.85\\
       M14	&0.85 & 0.80 & 0.78 & 0.84 & \textbf{0.87} & \textbf{0.87} \\
       M15	&0.84 & 0.81 & 0.83 & \textbf{0.86} & 0.84 & 0.84\\
       \midrule
       AVG	  &0.83 & 0.81 & 0.79 & 0.82 & 0.86 & \textbf{0.87}\\
       \bottomrule
    \end{tabular}
    \label{tab:ab_mp}
\end{table}

\begin{table}[!h]
    \centering
    \caption{Recall of different approaches to handle measurement issues. [a, b] denotes the label assigned to the missing entry in the hardcoding approach.}
    \begin{tabular}{c|c c c c | c | c}
       \toprule
       KQIs & [1,1] & [0,0] & [0,1] & [1,0] & Hard Loss & Soft Loss  \\
       \midrule
       M1	& 0.72 & 0.71 & 0.72 & 0.72 & 0.72 & \textbf{0.74} \\
       M2	& 0.70 & 0.69 & 0.68 & 0.70 & 0.74 & \textbf{0.75} \\
       M3 & 0.68 & 0.66 & 0.64 & 0.67 & 0.68 & \textbf{0.70}\\
       M4	&0.64 & 0.64 & 0.63 & 0.63 & 0.65 & \textbf{0.70}\\
       M5 &0.75 & 0.73 & 0.69 & 0.71 & 0.76 & \textbf{0.78} \\
       M6	&0.68 & 0.68 & 0.67 & 0.66 & 0.66 & \textbf{0.70} \\
       M7	&0.64 & 0.64 & 0.61 & 0.63 & 0.68 & \textbf{0.71}  \\
       M8	&0.80 & 0.78 & 0.78 & 0.77 & \textbf{0.82} & 0.80\\
       M9&0.70 & 0.72 & 0.70 & 0.66 & 0.74 & \textbf{0.79}\\
       M10	&0.66 & 0.66 & 0.68 & 0.65 & 0.71 & \textbf{0.74}\\
       M11	&0.81 & 0.81 & 0.78 & 0.81 & \textbf{0.83} & \textbf{0.83}\\
       M12	&0.70 & 0.70 & 0.70 & 0.69 & 0.71 & \textbf{0.75}\\
       M13	&0.73 & 0.75 & 0.74 & 0.71 & 0.76 & \textbf{0.78}\\
       M14	&0.51 & 0.51 & 0.51 & 0.49 & 0.54 & \textbf{0.64} \\
       M15	&0.53 & 0.55 & 0.55 & 0.55 & 0.58 & \textbf{0.62}\\
       \midrule
       AVG	  &0.69 & 0.70 & 0.68 & 0.66 & 0.72 & \textbf{0.77}\\
       \bottomrule
    \end{tabular}
    \label{tab:ab_mr}
\end{table}

\subsubsection{Handling Missing Measurements}
The main paper mentions that one key property of wafer sensor data is the frequent absence of fine-grained measurements, where each stage has an average of 3.6 measurements out of a total of 26 metrology metrics. To explore how the models can best tackle the issue, we carry out experiments to obtain the performance of several approaches: (1) hardcoding: we hard code the missing measurements with a specific label (\textit{e.g.}, [0, 0] for both the pass and fail labels); (2) Hard Loss Crop: For missing measurements, we do not assign the label and do not add their loss into the optimization; (3) Attention (Soft Loss): apart from the pass/fail, we predict attention for each assignment and leverage it as a weight for loss. Table~\ref{tab:ab_mp} and Table~\ref{tab:ab_mr} show the precision and recall for all three categories of the method, respectively. Overall, the performance of attention (soft loss) methods outperforms other approaches. On some measurements that have a smaller number of ground-truth labels, the precision of the attention method is much better, validating the usefulness of our design to handle the unbalanced inputs.

\subsubsection{Moving Window Size}

\section{Addtional Experimental Results in Generalizing Settings}
\label{sec:add_e}

\begin{table}[!t]
    \centering
    \caption{Comparisons of the state-of-the-art and our method in generalizability test (tool).}
    \begin{tabular}{c|c c | c c | c c}
       \toprule
       TOOL Splits & \multicolumn{2}{|c|}{Precision} & \multicolumn{2}{|c|}{Recall} & \multicolumn{2}{|c}{Variance} \\
       \midrule
       KQIs & SST & Ours & SST & Ours & SST & Ours \\
       \midrule
       M1	& 0.81 & \textbf{0.83} & 0.66 & \textbf{0.7} & 0.02 & 0.02 \\
       M2	& 0.84 & \textbf{0.85} & \textbf{0.72} & 0.71 & 0.01 & 0.02 \\
       M3 & 0.84 & \textbf{0.88} & \textbf{0.74} & 0.72 & 0.01 & 0.01\\
       M4	&0.82 & \textbf{0.85} & \textbf{0.69} & 0.67 & 0.02 & 0.03 \\
       M5	&\textbf{0.85} & 0.82 & 0.74 & \textbf{0.76} & 0.02 & 0.02 \\
       M6	& 0.85 & \textbf{0.87} & \textbf{0.75} & 0.72 & 0.01 & 0.01 \\
       M7	&0.81 & \textbf{0.83} & 0.71 & \textbf{0.74} & 0.03 & 0.05  \\
       M8	&\textbf{0.83} & \textbf{0.83} & 0.84 & \textbf{0.85} & 0.02 & 0.02\\
       M9	&0.7 & \textbf{0.75} & 0.7 & \textbf{0.74} & 0.11 & 0.09\\
       M10	&0.86 & \textbf{0.9} & \textbf{0.72} & 0.71 & 0.01 & 0.01\\
       M11	&\textbf{0.87} & 0.82 & 0.72 & \textbf{0.78} & 0.01 & 0.02\\
       M12	&0.79 & \textbf{0.81} & \textbf{0.76} & 0.74 & 0.08 & 0.14\\
       M13	&0.84 & \textbf{0.85} & 0.69 & \textbf{0.76} & 0.03 & 0.02\\
       M14	&\textbf{0.88} & 0.83 & 0.64 & \textbf{0.68} & 0.01 & 0.01 \\
       M15	&0.82 & \textbf{0.84} & \textbf{0.71} & 0.7 & 0.04 & 0.06\\
       \midrule
       AVG	  &0.82 & \textbf{0.84} & 0.73 & \textbf{0.74} & 0.05 & 0.09\\
       \bottomrule
    \end{tabular}
    \label{tab:gst}
\end{table}

\subsubsection{Generalizing Settings (Tool)}
In wafer manufacturing, there are a number of different products, which can differ a lot in water design and require distinct combinations of procedures conducted with diverse tools. To evaluate how our modular networks generalize over novel products/tools, we create two data splits with disjoint ``tool'' labels, respectively. Table~\ref{tab:gst} demonstrate the precision, recall, and variance of the state-of-the-art model (\textit{i.e.}, SST) and our proposed model in both splits. Results show that our proposed method outperforms the state-of-the-art in terms of all metrics over most fine-grained measurements, demonstrating better generalizability to different products and tools.

\begin{table}[!h]
    \centering
    \caption{Comparisons of the state-of-the-art and our method in generalizability test (cross-fac, tool).}
    \begin{tabular}{c|c c | c c | c c}
       \toprule
       Cross-Fac TOOL & \multicolumn{2}{|c|}{Precision} & \multicolumn{2}{|c|}{Recall} & \multicolumn{2}{|c}{Variance} \\
       \midrule
       KQIs & SST & Ours & SST & Ours & SST & Ours \\
       \midrule
       M1	& 0.81 & \textbf{0.85} & 0.64 & \textbf{0.69} & 0.02 & 0.02 \\
       M2	& 0.82 & \textbf{0.85} & \textbf{0.71} & \textbf{0.71} & 0.01 & 0.02 \\
       M3 & 0.83 & \textbf{0.84} & \textbf{0.72} & 0.71 & 0.01 & 0.01\\
       M4	& 0.82 & \textbf{0.84} & \textbf{0.66} & \textbf{0.66} & 0.02 & 0.03 \\
       M5 & 0.84 & \textbf{0.86} & 0.72 & \textbf{0.73} & 0.02 & 0.02 \\
       M6	& 0.85 & \textbf{0.91} & \textbf{0.73} & 0.72 & 0.01 & 0.01 \\
       M7	& 0.82 & \textbf{0.84} & 0.71 & \textbf{0.72} & 0.03 & 0.05  \\
       M8	&0.81 & \textbf{0.84} & 0.84 & \textbf{0.85} & 0.02 & 0.02\\
       M9	&0.71 & \textbf{0.72} & 0.70 & \textbf{0.74} & 0.14 & 0.16\\
       M10	&0.86 & \textbf{0.88} & 0.70 & \textbf{0.71} & 0.01 & 0.01\\
       M11	&0.85 & \textbf{0.88} & 0.69 & \textbf{0.76} & 0.01 & 0.02\\
       M12	&0.80 & \textbf{0.82} & 0.71 & \textbf{0.74} & 0.11 & 0.14\\
       M13	&0.84 & \textbf{0.85} & 0.69 & \textbf{0.75} & 0.03 & 0.02\\
       M14	&0.86 & \textbf{0.85} & 0.65 & \textbf{0.67} & 0.01 & 0.01 \\
       M15	&\textbf{0.82} & 0.81 & \textbf{0.70} & \textbf{0.70} & 0.04 & 0.06\\
       \midrule
       AVG	  &0.81 & \textbf{0.84} & 0.71 & \textbf{0.73} & 0.07 & 0.10\\
       \bottomrule
    \end{tabular}
    \label{tab:cft}
\end{table}
\subsubsection{Generalizing Settings (Cross Factory)}
Factory location also plays an important role in wafer manufacturing. It is relevant to quantities of environmental conditions (\textit{e.g.}, humidity, pressure) and hence influences the product quality significantly. To evaluate our modular network's capability to generalize those environmental changes, we created two more splits of products and tools in cross-factory settings. Specifically, apart from the labels of products/tools, data in the training/evaluation set come from different factories. Table~\ref{tab:cft} exhibit the performance of our proposed and state-of-the-art methods over a set of common fine-grained measurements. 
Compared with results in single-factory settings (Table~\ref{tab:gst}), both models gain a lower precision/recall and higher variance in most fine-grained measurements. It suggests that predicting fine-grained measurements is more challenging in cross-factory settings. \\
Despite the performance drop in cross-factory settings, our method still performs overwhelmingly better performance than the transformer-based method, validating the fact that our proposed network is more capable of generalizing to diverse settings by addressing the compositionality of processing stages. \\


\newpage


\vfill